\pdfoutput=1

\documentclass[11pt]{article}
\usepackage{hyperref}
\usepackage{booktabs}
\usepackage{multirow}
\usepackage[table,xcdraw]{xcolor}
\usepackage{array}
\usepackage{amsmath}
\usepackage{graphicx}
\usepackage{tcolorbox}
\usepackage{amssymb}
\graphicspath{{./png/}}
\DeclareMathOperator{\Retrieve}{Retrieve}
\DeclareMathOperator{\Rerank}{Rerank}
\DeclareMathOperator{\MR}{M_{R}}
\DeclareMathOperator{\MEI}{M_{EI}}
\DeclareMathOperator{\MKS}{M_{KS}}
\DeclareMathOperator{\MA}{M_A}

\usepackage[preprint]{acl}

\usepackage{times}
\usepackage{latexsym}

\usepackage[T1]{fontenc}

\usepackage[utf8]{inputenc}

\usepackage{microtype}

\usepackage{inconsolata}

\usepackage{graphicx}

\title{DualRAG: A Dual-Process Approach to Integrate Reasoning and Retrieval for Multi-Hop Question Answering}

\author{
  \textbf{Rong Cheng\textsuperscript{1}},
  \textbf{Jinyi Liu\textsuperscript{1}},
  \textbf{Yan Zheng\textsuperscript{1*}},
  \textbf{Fei Ni\textsuperscript{1}},\\
  \textbf{Jiazhen Du\textsuperscript{2}},
  \textbf{Hangyu Mao\textsuperscript{2}},
  \textbf{Fuzheng Zhang\textsuperscript{2}},
  \textbf{Bo Wang\textsuperscript{1}},
  \textbf{Jianye Hao\textsuperscript{1*}}
\\
  \textsuperscript{1}College of Intelligence and Computing, Tianjin University, \\
  \textsuperscript{2}Independent Researcher
}

\begin{document}
\maketitle

\begingroup
\renewcommand\thefootnote{*}
\footnotetext{
Corresponding to: 
Yan Zheng (yanzheng@tju.edu.cn) and
Jianye Hao (jianye.hao@tju.edu.cn)}
\endgroup

\begin{abstract}

Multi-Hop Question Answering~(MHQA) tasks permeate real-world applications, posing challenges in orchestrating multi-step reasoning across diverse knowledge domains. While existing approaches have been improved with iterative retrieval, they still struggle to identify and organize dynamic knowledge.
To address this, we propose DualRAG, a synergistic dual-process framework that seamlessly integrates reasoning and retrieval.
DualRAG operates through two tightly coupled processes: Reasoning-augmented Querying~(RaQ) and progressive Knowledge Aggregation~(pKA). They work in concert: as RaQ navigates the reasoning path and generates targeted queries, pKA ensures that newly acquired knowledge is systematically integrated to support coherent reasoning. This creates a virtuous cycle of knowledge enrichment and reasoning refinement.
Further, through targeted fine-tuning, DualRAG preserves its sophisticated reasoning and retrieval capabilities in smaller-scale models, demonstrating its versatility and core advantages across different scales.
Extensive experiments demonstrate that this dual-process approach substantially improves answer accuracy and coherence, approaching, and in some cases surpassing, the performance achieved with oracle knowledge access. These results establish DualRAG as a robust and efficient solution for complex multi-hop reasoning tasks.
We release DualRAG at \href{https://github.com/cbxgss/rag}{this https URL}.

\end{abstract}

\section{Introduction}

In recent years, large language models~(LLMs) have demonstrated exceptional capabilities in language understanding, generation, and reasoning tasks, even surpassing human performance on some benchmarks~\cite{model/gpt4, ar, vision-reason, vison-reason2}. 
However, despite the extensive knowledge these models acquire during training, they often exhibit hallucination issues and face limitations about their knowledge boundaries when dealing with domain-specific, dynamically evolving, or long-tail information~\cite{model/llm/hallucinated1, model/llm/hallucinated2, rag/later/trust}. 
To address these issues, researchers have introduced Retrieval-Augmented Generation~(RAG) systems, which enable LLMs to efficiently utilize external knowledge bases and search engines to obtain relevant information, thereby enhancing the accuracy and reliability of the generated content~\cite{rag/base, survey/kdd}.

Traditional RAG methods follow a retrieve-then-read paradigm, where documents are retrieved based on the original query and then used for answer generation~\cite{survey/rag}. 
While effective for simple tasks, such a fixed retrieval strategy struggles to adapt flexibly to the ever-changing knowledge demands when confronted with multi-hop questions.

\begin{figure}[t]
    \centering
    \includegraphics[width=0.5\textwidth]{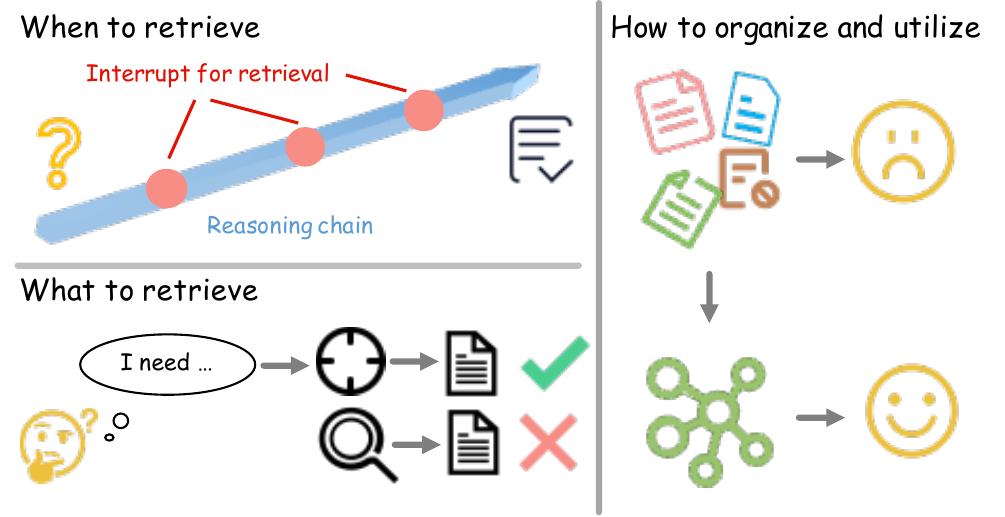}
    \caption{\textbf{Challenges in iterative RAG}, illustrating the evolving knowledge demands in multi-hop reasoning and the three core challenges.}
    \label{fig:challenge}
\end{figure}



When tackling complex multi-hop problems, the model encounters evolving knowledge demands as reasoning unfolds. This dynamic process gives rise to two primary challenges \textbf{\textit{when to retrieve}} and \textbf{\textit{what to retrieve}}. Although a series of iterative RAG systems have been proposed~\cite{rag/pipeline/ircot, rag/self/flare, rag/pipeline/planrag, rag/pipeline/genground, rag/pipeline/metarag}, most of these approaches lack the ability to proactively identify emerging knowledge gaps.This oversight often leads to interruptions in the LLM’s reasoning process frequently. Furthermore, the subsequent retrieval operations are not sufficiently targeted to bridge these gaps, which compromises the recall of relevant documents. Therefore, there is a pressing need for a method that can proactively accommodate shifting demands and effectively leverage retrieval tools to fill these knowledge gaps.

As retrieval demands increase, a third core challenge arises: \textbf{\textit{how to efficiently organize and utilize the retrieved information}}. Noise in retrieved documents, stemming from both the documents themselves and retrieval tools, is a common issue~\cite{rag/later/FoRAG, rag/later/inforag, rag/later/rgb-noise, rag/later/noise-nll-ft}.
In iterative RAG, noise accumulation can progressively interfere with the model’s understanding of available knowledge.
Moreover, poor organization of complex documents fragments knowledge, making it challenging for the model to construct a coherent reasoning chain~\cite{rag/rerank/lost-in-middle, rag/rerank/few-shot-in-translate}. Existing works have attempted to address this issue through re-ranking~\cite{rag/rerank/Selective-Context, rag/rerank/LLMLingua, rag/rerank/longllmlingua}, yet overlook the inherent associations among different documents.

To address the challenges above, we propose DualRAG, a novel iterative RAG framework with a dual-process architecture for efficiently coordinating reasoning and retrieval. DualRAG integrates two interdependent processes.
The primary process, Reasoning-augmented Querying (RaQ), acts as a diligent researcher, constructing reasoning chains, identifying knowledge gaps, and generating targeted queries when additional information is needed.
Meanwhile, the auxiliary process, progressive Knowledge Aggregation (pKA), serves as a dedicated assistant, filtering and organizing retrieved information into a coherent, evolving knowledge outline.
In this tightly coupled dual-process framework, the two processes continuously reinforce each other: RaQ provides explicit knowledge demands that guide pKA, while pKA continuously supplies a progressive knowledge outline to support RaQ's reasoning. This synergy enables DualRAG to dynamically adapt to evolving knowledge demands, efficiently bridge gaps, and maintain a noise-resilient foundation for complex multi-hop reasoning.
DualRAG is compatible with LLMs of various parameter scales, meaning its performance improves as the base models become more powerful. Given the lower computational cost of smaller models, we construct a specialized dataset and fine-tune them to enhance their capabilities, ensuring that DualRAG’s core advantages are preserved even in smaller-scale models.

To validate the effectiveness of our approach, we conducted extensive experiments on several multi-hop question-answering datasets.
Experimental results indicate that our framework achieves significant improvements across multiple key metrics compared to existing methods, demonstrating its superiority in handling complex reasoning tasks.

Our contributions can be summarized as follows: 
(1) We propose \textbf{DualRAG}, a dual-process framework where \textbf{\textit{RaQ}} guides reasoning and retrieval, while \textbf{\textit{pKA}} organizes retrieved knowledge to support inference.
(2) By identifying key entities, \textbf{\textit{RaQ}} dynamically generates targeted queries, while \textbf{\textit{pKA}} structures and integrates relevant information into a coherent knowledge outline, ensuring effective knowledge utilization;
(3) We develop a fine-tuned version of DualRAG, to enhance proficiency of LLMs in retrieval and generation, significantly reducing computational cost;
(4) {Extensive experiments} on multiple multi-hop question answering datasets validate the effectiveness and robustness of our approach.

\section{Related work}

\begin{figure*}[t]
    \centering
    \includegraphics[width=\textwidth]{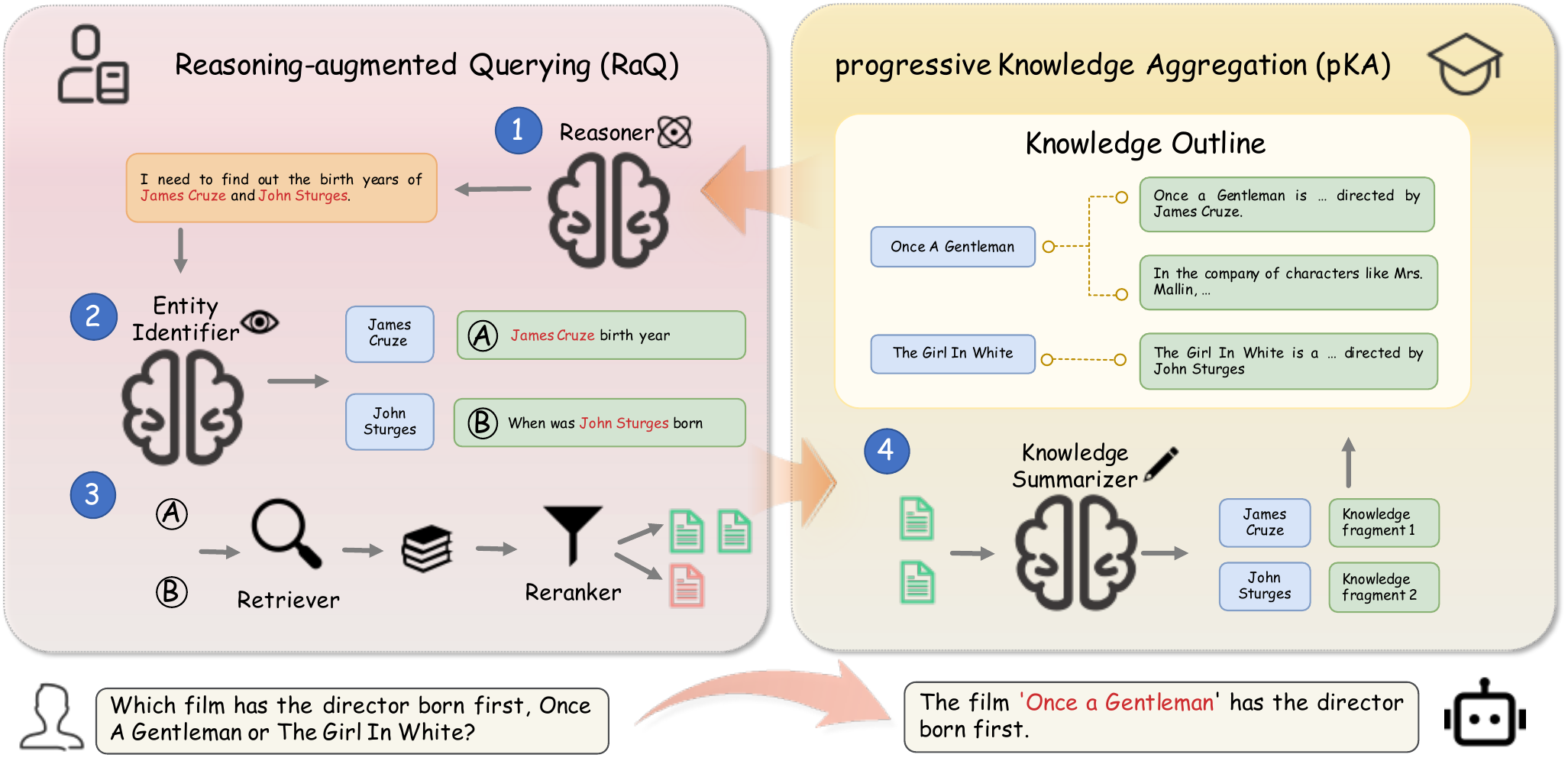}
    \caption{\textbf{Overview of DualRAG}, an iterative RAG framework for MHQA that combines Active Reasoning and Querying with Progressive Knowledge Aggregation.}
    \label{fig:main}
\end{figure*}

The development of Retrieval-Augmented Generation~(RAG) methods has progressed from traditional single-round retrieval to iterative RAG with multi-turn retrieval~\cite{survey/rag}, better adapting to the demands of complex reasoning tasks.

\subsection{RAG}

The earliest Retrieval-Augmented Generation (RAG) methods adopted the retrieve-then-read paradigm. Initially, a retriever fetches relevant documents from a corpus, and then a generative model produces an answer based on these documents. Common retrieval methods include sparse retrieval (e.g., BM25~\cite{model/bm25}), dense retrieval (e.g., E5~\cite{model/e5}, DPR~\cite{rag/retreive/dense}), and search engines like Bing and Google.

To enhance retrieval accuracy, researchers have proposed various optimization strategies. D2LLM~\cite{model/retreive/d2llm} transfers the computationally expensive cross-encoder capabilities to a more efficient bi-encoder model. MRAG~\cite{rag/retreive/mrag} introduces multi-head attention mechanisms to encode documents into multiple vectors, capturing semantic information more comprehensively. Additionally, some studies utilize reranking techniques to filter retrieval results, ensuring that only the most relevant knowledge is retained \cite{model/rerank, model/rerank/rankrag}. LongLLMLingua~\cite{rag/rerank/longllmlingua} further optimizes the document ranking order.

Meanwhile, some research has explored leveraging the inherent knowledge capabilities of large models to enhance the adaptability of retrieval strategies, thereby reducing unnecessary external queries. For instance, SKR~\cite{rag/self/skr} assesses the complexity of the the given question by comparing it with similar past questions. FLARE~\cite{rag/self/flare} and DRAGIN~\cite{rag/self/dragin} trigger external retrieval when the model's output logits indicate uncertainty.

Given that the effectiveness of RAG systems heavily depends on query quality, many studies focus on optimizing query formulation to enhance retrieval recall. Methods like HyDE~\cite{rag/rewrite/hyde} and Query2doc~\cite{rag/rewrite/query2doc} generate a pseudo-document based on the question, which is then used for retrieval. RRR~\cite{rag/rewrite/rrr} introduces a rewrite-retrieve-read paradigm and fine-tunes the rewrite model using PPO.

Unlike the previous approaches, our method acknowledges that knowledge needs evolve dynamically throughout the reasoning process and the acquisition of new knowledge, tackling complex multi-hop problems through an iterative approach.

\subsection{Iterative RAG}

Although early RAG methods achieved some success in knowledge retrieval, their limitations in complex reasoning tasks have led to the development of Iterative Retrieval-Augmented Generation (iterative RAG) frameworks. These frameworks enable models to retrieve external knowledge multiple times throughout the reasoning process, gradually constructing a complete reasoning path for complex reasoning tasks. IRCoT~\cite{rag/pipeline/ircot}, Iter-RetGen~\cite{rag/pipeline/iter-retgen}, In Contenx-RALM~\cite{rag/pipeline/in-context-ralm} leverage previously generated content from LLMs to trigger retrieval at predefined intervals. FLARE~\cite{rag/self/flare}, DRAGIN~\cite{rag/self/dragin} utilize model output logits as signals as trigger retrieval signals. PlanRAG~\cite{rag/pipeline/planrag}, Plan$\times$RAG~\cite{rag/pipeline/planxrag}, GenGround~\cite{rag/pipeline/genground} decompose the original question into sub-questions, retrieve information separately, and synthesize a final answer. SlimPLM~\cite{rag/self/slimrag}, MetaRAG~\cite{rag/pipeline/metarag} generate heuristic answers first, then refine them through retrieval. Self-RAG~\cite{rag/self/selfrag} introduces special tokens during training, allowing models to control the retrieval directly through these tokens.

Unlike existing methods, our approach proactively identifies knowledge needs during reasoning, retrieves the relevant information, and organizes the knowledge into a coherent knowledge base, enabling more effective multi-hop reasoning.

\section{DualRAG}

Existing methods struggle to dynamically identify evolving knowledge demands during reasoning and to effectively organize retrieved information, weakening retrieval-enhanced reasoning.
To address this, we introduce DualRAG, which tightly integrates retrieval and reasoning through two interdependent processes: Reasoning-augmented Querying~(RaQ) and progressive Knowledge Aggregation~(pKA).
Furthermore, DualRAG is compatible with LLMs of various scales—stronger models further enhance its performance, while smaller ones may see a slight drop in performance. To mitigate this, we fine-tune them on a specialized dataset, ensuring effectiveness with minimal performance loss and enabling a smooth transition to smaller models, thereby reducing computational costs.

\subsection{Framework of DualRAG}

To address the aforementioned challenges, we propose a dual-process closed-loop framework centered around two tightly interconnected core processes: Reasoning-augmented Querying~(RaQ) and progressive Knowledge Aggregation~(pKA). These two processes operate in a synergistic and iterative manner, continuously refining knowledge acquisition and reasoning.

To begin with, we formally define the RAG task to enhance clarity, as follows: Given a user question $x$ and a large-scale document corpus $\mathcal{D} = \{d_i\}_{i=1}^{N}$, the objective of a RAG system is to generate an accurate answer $\hat{a}$ by retrieving and leveraging relevant documents from $\mathcal{D}$.

Our framework is illustrated in Figure~\ref{fig:main}.
RaQ acts as a diligent researcher, reasoning over the progressive knowledge outline $K$ maintained by pKA while dynamically identifying missing information and generating targeted retrieval queries, thus ensures a continuous flow of knowledge demands and potentially relevant documents $D$ into pKA.
Meanwhile, pKA serves as a dedicated assistant, integrating retrieved documents into a progressive knowledge outline $K$, which continuously supports RaQ’s reasoning.
This closed-loop interaction enables the system to iteratively refine both the reasoning process and the external knowledge integration. Formally, for the $t$-th iteration, this iterative collaboration can be expressed as follows:
\begin{align}
    r^t, D^t &= \mathrm{RaQ}(K^{t-1}, x, R^{t-1}) \label{eq:raq} \\
    K^t &= \mathrm{pKA}(K^{t-1}, D^t) \label{eq:pka}
\end{align}
where $r^t$ represents the reasoning outcome at step $t$, $D^t$ denotes the retrieved documents and $K^t$ is the updated knowledge outline. The accumulated reasoning history is captured as $R^{t-1} = \left\{ r^1, r^2, \cdots, r^{t-1}  \right\}$.

In the following sections, we elaborate on the mechanisms of RaQ and pKA and their synergy in the dual-process framework.

\subsubsection{Reasoning-augmented Querying~(RaQ)}

The RaQ process aims to dynamically identify emerging knowledge demands during reasoning and formulates queries accordingly. To achieve this, we guide LLMs to assess knowledge gaps and generate queries to retrieve relevant information to expand the knowledge closure. This process is facilitated by two collaborative components: the \textbf{Reasoner} and \textbf{Entity Identifier}.

\paragraph{Reasoner} The Reasoner advances reasoning based on the current knowledge outline $K^{t-1}$ maintained by pKA and the previous reasoning history $R^{t-1}$. It also determines whether retrieval is necessary by identifying knowledge gaps. Formally, this process can be expressed as:
\begin{equation}
    r^t, f^t = \MR(K^{t-1},\, x,\, r^{t-1})
\label{eq:reasoner}
\end{equation}
where $f^t$ denotes the retrieval trigger flag. If $f^t = \text{False}$, it indicates that no additional knowledge is required for reasoning, the final answer is then generated using the aggregated external knowledge $K^T$ and the complete reasoning history $R^T$:
\begin{equation}
    \hat{a} = \MA(K^T, x, R^T)
\label{eq:answer}
\end{equation}
where $K^T$ and $R^T$ denote the final knowledge outline and reasoning history, respectively.

\paragraph{Entity Identifier} Once the Reasoner detects a knowledge gap ($f^t = \text{True}$), retrieval is triggered to obtain relevant information. Prior studies have demonstrated that query rewriting significantly improves retrieval recall~\cite{rag/rewrite/rrr, rag/rewrite/query2doc}. Knowledge is often centered around entities, which serve as core carriers of diverse related information. Thus, we guide LLMs to identify key entities or concepts relevant to the current knowledge demand. This serves two purposes. First, it enables the generation of multiple queries, each capturing different aspects of the entity’s knowledge. Second, in the \S~\ref{sec: pKA}, pKA will organize knowledge around these entities. Formally, this process is represented as:
\begin{equation}
    E^t, \{ Q^t(e) \}_{e \in E^t} = \MEI(K^{t-1},\, x,\, r^t)
\label{eq:query}
\end{equation}
where $E^t = \left\{ e_1^t, e_2^t, \cdots, e_K^t \right\}$ denotes the set of identified key entities, $\{ Q^t(e) \}_{e \in E^t}$ denotes the set of queries associated with each entity $e \in E^t$. To maintain consistency in entity identification across reasoning steps, the Entity Identifier also links current key entities to synonymous counterparts from previous iterations.

Subsequently, each query $q \in Q^t(e)$ retrieves relevant documents from the corpus $\mathcal{D}$:
\begin{equation}
    \hat{D}_{e,q} = \Retrieve(q) \quad \text{for each } q \in Q^t(e)
\label{eq:retriever}
\end{equation}
To enable efficient learning and integration of retrieval results by pKA, we first Group documents by entity and apply a reranker model for initial filtering. This process is formalized as follows:
\begin{align}
    \hat{D}_e &= \bigcup_{q \in Q^t(e)}\hat{D}_{e,q} \label{eq:doc_join} \\
    D_e &= \Rerank(e,\, \hat{D}_e) \label{eq:reranker}
\end{align}

Through the collaboration of Reasoner and Entity Identifier, the RaQ process dynamically identifies knowledge demands and retrieves entity-structured documents while advancing the reasoning chain. This ensures a continuous document flow to pKA.

\subsubsection{Progressive Knowledge Aggregation}
\label{sec: pKA}

The pKA process aims to maintain a progressive knowledge outline, serving as external knowledge support $K^t$ for the RaQ process. Studies have shown that retrieval results often contain noise~\cite{rag/later/rgb-noise, rag/later/noise-nll-ft}, and the sequential organization of documents greatly affects LLM output~\cite{rag/rerank/lost-in-middle, rag/rerank/longllmlingua}. To address these issues, we propose knowledge-demand-driven summarization and an entity-based knowledge organization structure.

\paragraph{Knowledge Summarizer} Although the RaQ process has applied initial document-level filtering, substantial noise remains in individual documents. We guide LLMs to summarize retrieved results $D^t = \{ D_e^* \}_{e \in E^t}$ in a knowledge-demand-driven manner. This ensures that only essential knowledge is retained while reducing noise and redundancy. Formally, for each key entity $e$, the summarized knowledge fragment $k_e$ is obtained as follows:
\begin{equation}
    k_e = \MKS(x,\, R^t,\, e,\, Q^t(e), D_e)
\label{eq:summarizer}
\end{equation}

\paragraph{Progressive Knowledge Outline} In the previous step, we summarized the retrieved documents. Next, newly acquired knowledge fragments $k_e$ are merged with previously accumulated knowledge for each entity, integrating them into the Knowledge Outline $K$:
\begin{equation}
    K^t(e) = K^{t-1}(e) \cup \{ k_e \}
\label{eq:knowleage_join}
\end{equation}
Initially, the knowledge outline is empty, $K^0 = \emptyset$. This structured representation links knowledge to specific entities, allowing the Reasoner in the RaQ process to better utilize the available knowledge.

Through the Knowledge Summarizer and the progressive maintenance of the Knowledge Outline, the pKA process effectively filters, structures, and continuously integrates retrieved information. This creates a dynamically evolving knowledge foundation for reasoning in the RaQ process.

\subsection{Fine-Tuning for Compact Models}
\label{sec:finetune}

DualRAG is compatible with LLMs of various parameter scales, where stronger models yield better performance but also incur disproportionately higher computational costs~\cite{scaling-laws}. Consequently, while smaller models significantly reduce computational costs, their performance may degrade due to limited capacity.

To mitigate this, we construct a specialized dataset to fine-tune smaller models and enhance their capabilities. Using Qwen2.5-72B-Instruct~\cite{model/qwen2.5}, we apply DualRAG to generate 5,000 complete reasoning trajectories from the training sets of HotpotQA~\cite{benchmark/hotpotqa}, 2WikiMultihopQA~\cite{benchmark/2wikimultihop}, and MuSiQue~\cite{benchmark/musique}. We then filter these trajectories, retaining only those that lead to correct answers. From the filtered data, we derive a targeted initial training dataset aimed at improving three key components: the Reasoner, the Entity Identifier, and the Knowledge Summarizer.

Subsequently, we conduct a detailed analysis of the dataset and perform the following post-processing. 
(1) Entity Identifier: Even the large teacher model sometimes generates redundant or ineffective queries. To refine these, we employ a cross-encoder model to compute similarity between identified entities, perform entity linking, and remove redundant entities. This reduces unnecessary retrieval operations and lowers context overhead.
(2) Knowledge Summarizer: Although the large model produces accurate summaries in most cases, it occasionally omits key information in multi-hop reasoning scenarios due to difficulty in identifying implicit connections between retrieved documents and the question. To remedy this, we compare model-generated summaries with ground-truth documents to ensure latent information is preserved, thereby enhancing knowledge integration integrity and reasoning accuracy.

Table~\ref{sft-statistic} presents dataset statistics. Fine-tuning on this dataset enables our method to transition to more compact, computationally efficient models while preserving its core advantages.

\begin{table}[t]
\centering
\setlength{\tabcolsep}{4.5pt}
\small
\begin{tabular}{lrrr}
    \toprule
    \textbf{Capability} & \textbf{Raw} & \textbf{Processed} & \textbf{Count} \\
    \midrule
    Reasoner & 33,342 & 0 & 33,342 \\
    Entity Identifier & 4,951 & 17,158 & 22,109\\
    Knowledge Summarizer & 22,058 & 9,559 & 31,617\\
    \midrule
    Total & 60351 & 26717 & 87,068 \\
    \bottomrule
\end{tabular}
\caption{Statistics of the train dataset}
\label{sft-statistic}
\end{table}

\begin{table*}[t]
\centering
\small
\setlength{\tabcolsep}{5.5pt}
\renewcommand{\arraystretch}{1.1}
\begin{tabular}{lcccccccccccc}
\toprule
\multirow{2}{*}{\textbf{Methods}} 
    & \multicolumn{4}{c}{\textbf{HotpotQA}} 
    & \multicolumn{4}{c}{\textbf{2Wikimultihopqa}} 
    & \multicolumn{4}{c}{\textbf{MuSiQue}} \\
\cmidrule(lr){2-5} \cmidrule(lr){6-9} \cmidrule(lr){10-13}
    & \textbf{$\text{Acc}^{\dagger}$} & \textbf{EM} & \textbf{Acc} & \textbf{F1}
    & \textbf{$\text{Acc}^{\dagger}$} & \textbf{EM} & \textbf{Acc} & \textbf{F1}
    & \textbf{$\text{Acc}^{\dagger}$} & \textbf{EM} & \textbf{Acc} & \textbf{F1} \\
\midrule
\rowcolor[gray]{0.9} \textit{Base LLM} & \multicolumn{12}{c}{\textit{Qwen2.5-72B-Instruct}} \\
Direct
    & 42.1 & 26.0 & 29.1 & 36.4
    & 33.0 & 28.6 & 31.0 & 35.0
    & 19.4 & 8.3 & 11.8 & 17.5 \\
NativeRAG~\cite{rag/base}
    & 69.3 & 46.4 & 54.3 & 60.3
    & 50.5 & 40.6 & 46.5 & 48.1
    & 39.4 & 23.0 & 30.2 & 33.6 \\
IRCOT~\cite{rag/pipeline/ircot}
    & \underline{79.4} & \textbf{52.5} & \underline{69.7} & \textbf{67.4}
    & \underline{77.2} & 56.6 & \underline{83.6} & 67.5
    & \underline{58.3} & \underline{34.3} & 51.0 & 48.1 \\
MetaRAG~\cite{rag/pipeline/metarag}
    & 74.3 & 53.4 & 58.8 & \underline{66.9}
    & 63.1 & 54.2 & 59.7 & 61.2
    & 57.4 & 39.3 & 47.5 & \underline{51.7} \\
GenGround~\cite{rag/pipeline/genground}
    & 78.7 & 46.4 & 64.8 & 61.8
    & 76.3 & \underline{57.3} & 78.1 & \underline{70.3}
    & 54.8 & 28.8 & \underline{53.0} & 43.0 \\
\textit{Oracle}
    & \textit{89.3} & \textit{64.3} & \textit{72.4} & \textit{79.6}
    & \textit{88.0} & \textit{76.1} & \textit{85.6} & \textit{83.5}
    & \textit{75.6} & \textit{56.6} & \textit{68.0} & \textit{69.1} \\
\midrule
\textbf{DualRAG}
    & \textbf{79.7} & \underline{49.7} & \textbf{70.0} & 65.7
    & \textbf{84.8} & \textbf{65.6} & \textbf{85.0} & \textbf{77.3}
    & \textbf{70.1} & \textbf{40.8} & \textbf{66.3} & \textbf{56.3} \\
\midrule
\midrule
\rowcolor[gray]{0.9} \textit{Base LLM} & \multicolumn{12}{c}{\textit{Qwen2.5-7B-Instruct}} \\
Direct
    & 27.3 & 17.4 & 19.1 & 25.0
    & 26.1 & 23.5 & 24.3 & 28.2
    & 10.3 & 4.6 & 6.4 & 9.9 \\
NativeRAG~\cite{rag/base}
    & 57.9 & 38.5 & 43.9 & 49.5
    & 32.1 & 26.8 & 29.2 & 30.9
    & 22.0 & 12.6 & 15.7 & 20.2 \\
IRCOT~\cite{rag/pipeline/ircot}
    & 68.3 & 38.9 & 60.6 & 52.8
    & 53.5 & 38.2 & 62.6 & 48.8
    & 34.0 & 13.9 & 29.6 & 24.6 \\
MetaRAG~\cite{rag/pipeline/metarag}
    & 62.9 & \underline{44.4} & 49.4 & 56.8
    & 46.2 & 40.1 & 43.2 & 46.5
    & 39.2 & 28.3 & 33.6 & 37.9 \\
GenGround~\cite{rag/pipeline/genground}
    & 66.4 & 36.0 & 58.2 & 50.0
    & 47.1 & 37.4 & 57.0 & 47.5
    & 42.3 & 17.5 & 38.8 & 30.6 \\
\textit{Oracle}
    & \textit{76.4} & \textit{55.1} & \textit{60.1} & \textit{67.6}
    & \textit{63.3} & \textit{53.7} & \textit{60.1} & \textit{59.3}
    & \textit{61.6} & \textit{38.6} & \textit{45.3} & \textit{48.1} \\
\midrule
\textbf{DualRAG}
    & \underline{72.2} & 43.4 & \underline{64.1} & \underline{58.6}
    & \underline{68.6} & \underline{53.2} & \underline{75.8} & \underline{64.4}
    & \underline{56.3} & \underline{29.9} & \underline{52.5} & \underline{44.9} \\
\textbf{DualRAG-FT}
    & \textbf{76.3} & \textbf{45.6} & \textbf{64.8} & \textbf{61.6}
    & \textbf{81.2} & \textbf{61.8} & \textbf{82.0} & \textbf{74.6}
    & \textbf{58.6} & \textbf{32.7} & \textbf{52.8} & \textbf{46.5} \\
\bottomrule
\end{tabular}
\caption{Evaluation results on three MHQA datasets, using Qwen2.5-72B-Instruct and Qwen2.5-7B-Instruct as the base LLMs. \textbf{Bold} indicates the best performance and \underline{underline} denotes the second-best. \textit{Oracle} represents an upper bound where the LLM receives key information directly from ground-truth relevant documents, bypassing retrieval.}
\label{tab:exp-main}
\end{table*}

\section{Experiment}

\subsection{Datasets and Metrics}

We evaluate our method on three open-domain multi-hop question answering datasets: HotpotQA~\cite{benchmark/hotpotqa}, 2WikiMultihopQA~\cite{benchmark/2wikimultihop}, and MuSiQue~\cite{benchmark/musique}. For HotpotQA, we use its official Wikipedia corpus as the retrieval database. Since 2WikiMultihopQA and MuSiQue do not provide an official corpus, we construct the retrieval database by merging all supporting and non-supporting passages from each dataset. Due to computational constraints, we randomly sample 1,000 questions from the dev or test subsets of each dataset for evaluation. More details can be found in Appendix~\ref{appaendix:datasets}.

Regarding evaluation metrics, we use the following standard measures to assess the quality of generated responses: \textbf{Exact Match (EM)}, which measures the degree of exact matching between the generated answer and the ground truth; \textbf{Acc}, which measures whether the generated answer adequately captures the key content of the ground truth; and \textbf{Token-level F1}, which evaluates the token-level similarity between the generated and ground truth answers. \textbf{$\text{Acc}^{\dagger}$}, assesses correctness using a LLM, with details provided in Appendix~\ref{appendix:metric}.

\subsection{Baselines}

We first consider a non-retrieval baseline: (1) \textbf{Direct}, which generates answers without retrieving external knowledge.

Next, we include a standard RAG method: (2) \textbf{NativeRAG~\cite{rag/base}}, which follows a retrieve-then-read paradigm.

Finally, we consider iterative RAG methods, including (3) \textbf{IRCOT~\cite{rag/pipeline/ircot}}, which retrieves information at predefined intervals during reasoning process; (4) \textbf{MetaRAG~\cite{rag/pipeline/metarag}}, which
refines the initial answer through multiple rounds of retrieval; (5) \textbf{GenGround~\cite{rag/pipeline/genground}}, which decomposes the original question into multiple sub-questions for retrieval.

\subsection{Implementation Details}

All iterative RAG methods are set with maximum of 5 iteration steps. Training parameters are provided in Appendix~\ref{appdendix:training}.

\paragraph{Retrieval Module} We employ faiss-gpu to build an efficient vector index and use bge-small-en-v1.5~\cite{model/embedding} as the document encoder model. To enhance retrieval quality, we use bge-reranker-v2-m3~\cite{model/rerank} as the reranker and deploy an online retrieval service based on fastapi. To ensure fair comparisons across all baseline methods, we incorporate the reranker to filter retrieval results. During retrieval, we first recall the top-50 documents and then use the reranker to further refine the selection to the top-10 documents for model reasoning.

\paragraph{Language Models} We conduct experiments on both Qwen-2.5-72B-Instruct and Qwen-2.5-7B-Instruct~\cite{model/qwen2.5} models and utilize vllm~\cite{model/vllm} for efficient inference deployment.

\subsection{Experimental Results}

\begin{figure}[t]
    \centering
    \includegraphics[width=0.45\textwidth]{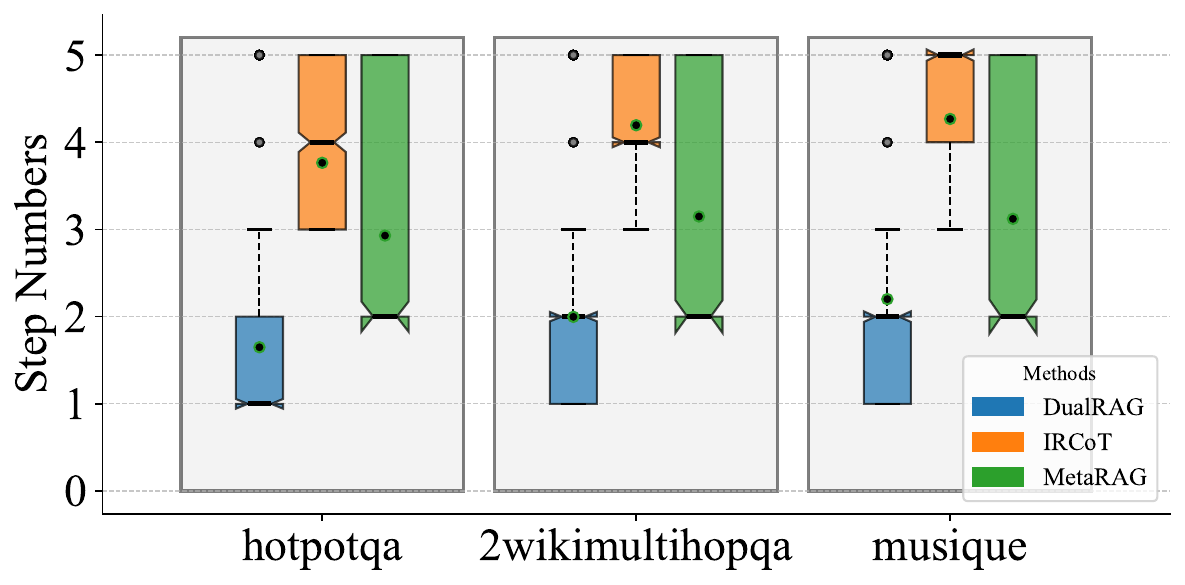}
    \caption{The \textbf{distribution of the average number of iterations} per question for each method. Note that GenGround does not specify a clear termination criterion and always iterates up to the preset maximum limit. Therefore, its iteration count distribution is not included in the figure.}
    \label{fig:exp/steps}
\end{figure}

The evaluation results are shown in Table~\ref{tab:exp-main}. Overall, these results demonstrate that DualRAG achieves substantial performance improvements across a variety of datasets. We make the following key observations:

(1) Incorporating external knowledge consistently outperforms relying solely on model parameters, confirming the critical role of retrieval in addressing knowledge gaps in multi-hop tasks.

(2) Iterative retrieval allows the system to progressively expand its knowledge closure by capturing the evolving information needs during reasoning. This results in a more comprehensive support for inference compared to single-round retrieval.

(3) Among iterative RAG methods, DualRAG stands out due to its dual closed-loop design, which tightly integrates active reasoning with dynamic external information aggregation. This mechanism for generating targeted queries and structurally organizing the retrieved data leads to more accurate and coherent answers.

(4) DualRAG-FT achieves significant performance improvements through fine-tuning, allowing a reduction in computational costs with smaller models while maintaining strong performance.

(5) As illustrated in Figure~\ref{fig:exp/steps}, DualRAG significantly reduces the number of iterative retrieval steps, thereby minimizing interruptions to the LLM’s reasoning process. Notably, our framework minimizes retrievals to the lowest feasible level, demonstrating that DualRAG proactively identifies and fulfills knowledge needs rather than relying on repeated trial-and-error. A detailed analysis is provided in Appendix~\ref{app:scaling}.

\subsection{Ablation Study}

We conduct a series of ablation studies to assess the contributions of individual components to our framework and to evaluate the effectiveness of our fine-tuning. Our analyses cover both the framework itself and the impact of fine-tuning on each module.

\begin{table}[t]
\centering
\small
\setlength{\tabcolsep}{4.5pt}
\begin{tabular}{lcccccc}
\toprule
\multirow{2}{*}{\textbf{Methods}} 
    & \multicolumn{2}{c}{\textbf{HQA}} 
    & \multicolumn{2}{c}{\textbf{WQA}} 
    & \multicolumn{2}{c}{\textbf{MQA}} \\
\cmidrule(lr){2-3} \cmidrule(lr){4-5} \cmidrule(lr){6-7}
    & \textbf{Acc} & \textbf{F1}
    & \textbf{Acc} & \textbf{F1}
    & \textbf{Acc} & \textbf{F1} \\
\midrule
\textbf{DualRAG}
    & 70.0 & 65.7
    & 85.0 & 77.3
    & 66.3 & 56.3 \\
- w/o R
    & 68.7 & 64.2
    & 84.2 & 73.2
    & 56.1 & 48.6 \\
- w/o EI
    & 69.2 & 65.3
    & 83.7 & 72.5
    & 54.2 & 48.4 \\
- w/o KO
    & 69.5 & 64.0
    & 79.4 & 72.8
    & 58.1 & 49.8 \\
\bottomrule
\end{tabular}
\caption{Ablation Study on DualRAG Framework Components using Qwen2.5-72B-Instruct.}
\label{tab:exp-abl-framework}
\end{table}

\paragraph{Ablation on the Framework}
As shown in Table~\ref{tab:exp-abl-framework}, we examine the framework by incrementally removing key components. First, we disable Reasoner's active exploration for missing knowledge (\textit{w/o} R), mirroring the retrieval behavior in IRCoT. The performance drop underscores that detecting missing information during reasoning is critical for generating effective queries. Next, we remove Entity Identifier, using Reasoner’s output as the query (\textit{w/o} Ei). This modification leads to a notable decline in performance, indicating that generating queries tailored to specific knowledge needs is essential. Finally, we eliminate the Knowledge Outline mechanism, feeding the model unstructured retrieval results (\textit{w/o} KO). The corresponding performance drop underscores the importance of organizing and structuring retrieved information for subsequent reasoning.

\begin{table}[t]
\centering
\small
\setlength{\tabcolsep}{4pt}
\begin{tabular}{lcccccc}
\toprule
\multirow{2}{*}{\textbf{Methods}} 
    & \multicolumn{2}{c}{\textbf{HQA}} 
    & \multicolumn{2}{c}{\textbf{WQA}} 
    & \multicolumn{2}{c}{\textbf{MQA}} \\
\cmidrule(lr){2-3} \cmidrule(lr){4-5} \cmidrule(lr){6-7}
    & \textbf{Acc} & \textbf{F1}
    & \textbf{Acc} & \textbf{F1}
    & \textbf{Acc} & \textbf{F1} \\
\midrule
\textbf{DualRAG-FT}
    & 64.8 & 61.6
    & 82.0 & 74.6
    & 52.8 & 46.5 \\
- w/o R-FT
    & 63.4 & 59.9
    & 77.6 & 70.2
    & 51.0 & 46.4 \\
- w/o EI-FT
    & 65.7 & 61.7
    & 79.5 & 73.6
    & 51.7 & 46.5 \\
- w/o KS-FT
    & 60.7 & 58.5
    & 76.4 & 66.4
    & 51.9 & 46.2 \\
- w/o FT
    & 64.1 & 58.6
    & 75.8 & 64.4
    & 52.5 & 44.9 \\
\bottomrule
\end{tabular}
\caption{Ablation Study on the fine-tune for DualRAG Using Qwen2.5-7B-Instruct.}
\label{tab:exp-abl-sft}
\end{table}

\begin{figure}[t]
    \centering
    \includegraphics[width=0.45\textwidth]{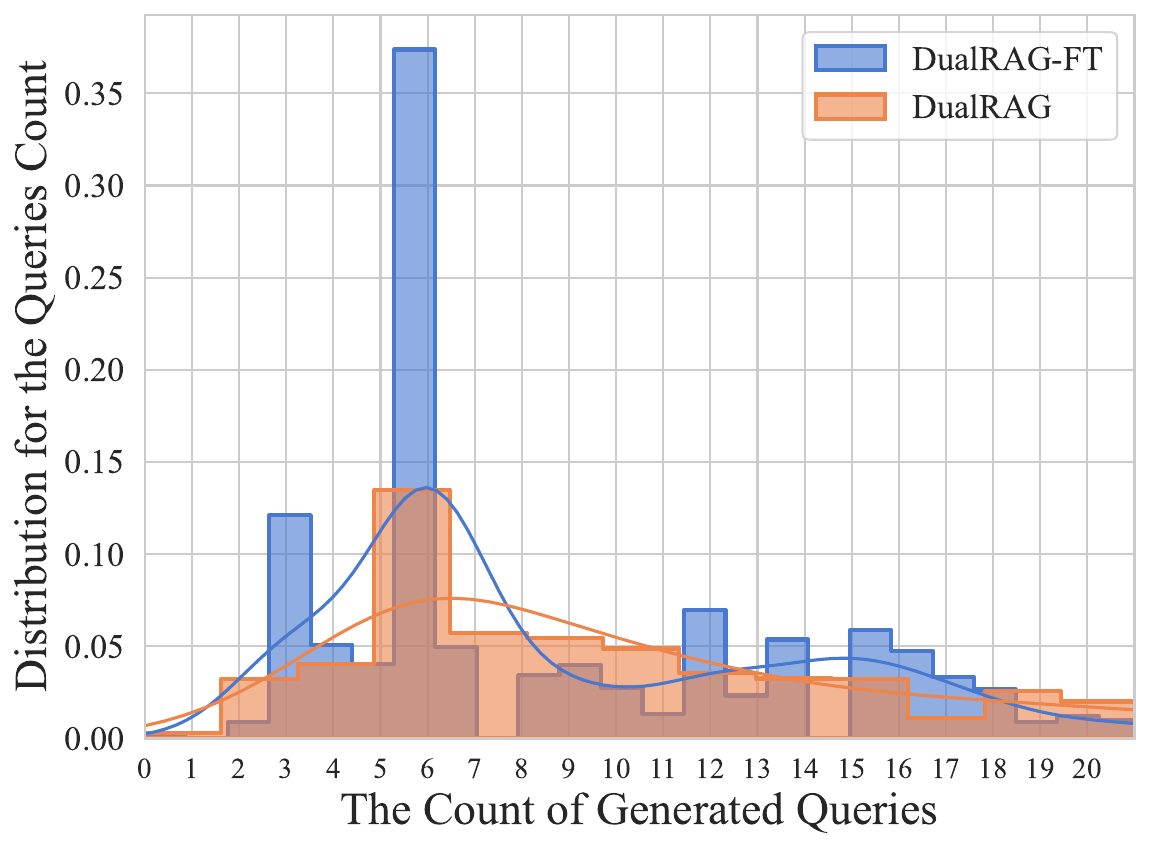}
    \caption{Comparison of the count of queries generated per question by DualRAG and DualRAG-FT. The DualRAG-FT produces fewer redundant queries, thereby reducing unnecessary retrieval calls.}
    \label{fig:exp/sft-EI}
\end{figure}

\paragraph{Ablation on the SFT}
As shown in Table~\ref{tab:exp-abl-sft}, we further investigate the impact of fine-tuning by comparing DualRAG-FT with its variants without fine-tuning on individual components. In the \textit{w/o} R-FT, \textit{w/o} EI-FT, and \textit{w/o} KS-FT settings, the Reasoner, Entity Identifier, and Knowledge Summarizer are replaced with their counterparts without being fine-tuned on data from \S~\ref{sec:finetune}. In the \textbf{w/o} FT setting, no component is fine-tuned. The experimental results indicate that the fine-tuned Reasoner and Knowledge Summarizer significantly enhance the small model’s ability to utilize retrieved information. Moreover, as shown in Figure~\ref{fig:exp/sft-EI}, the fine-tuned Entity Identifier effectively reduces redundant query generation, thereby decreasing unnecessary retrieval calls.

Overall, these ablation studies highlight the critical role of fine-tuning all three components in improving the small model's performance.

\subsection{Other QA Tasks}

Beyond primary MHQA datasets, we also evaluate our approach on other QA datasets, as shown in Tables~\ref{tab:exp-single-hop} and \ref{tab:exp-long-format}. The results show strong performance across QA tasks, highlighting notable advantages over the conventional retrieve-then-read paradigm in RAG~\cite{survey/rag, survey/flashrag}.

\begin{table}[t]
\centering
\small
\setlength{\tabcolsep}{4pt}
\begin{tabular}{lcccccc}
\toprule
\multirow{2}{*}{\textbf{Methods}}
    & \multicolumn{3}{c}{\textbf{NQ}} 
    & \multicolumn{3}{c}{\textbf{PopQA}} \\
\cmidrule(lr){2-4} \cmidrule(lr){5-7}
    & \textbf{$\text{Acc}^{\dagger}$} & \textbf{Acc} & \textbf{F1}
    & \textbf{$\text{Acc}^{\dagger}$} & \textbf{Acc} & \textbf{F1} \\
\midrule
Direct
    & 42.1 & 37.9 & 38.8
    & 34.9 & 30.9 & 26.3 \\
Native
    & 60.8 & 48.5 & 44.5
    & 77.1 & 74.7 & \textbf{57.9} \\
IRCoT
    & 64.7 & 52.9 & 44.8
    & 70.3 & 70.0 & 48.1 \\
MetaRAG
    & 59.8 & 48.1 & \textbf{48.3}
    & 76.9 & 72.8 & \underline{58.3} \\
GenGround
    & \underline{68.1} & \underline{57.2} & 44.5
    & 73.9 & 72.9 & 49.6 \\
\textbf{DualRAG}
    & \textbf{68.2} & \textbf{57.9} & \underline{45.0}
    & \textbf{78.9} & \textbf{79.0} & 52.8 \\
\bottomrule
\end{tabular}
\caption{Evaluation results on simple single-hop question answering datasets, using Qwen2.5-72B-Instruct.}
\label{tab:exp-single-hop}
\end{table}

\begin{table}[t]
\centering
\small
\setlength{\tabcolsep}{4pt}
\begin{tabular}{lcccccc}
\toprule
\multirow{2}{*}{\textbf{Methods}} 
    & \multicolumn{2}{c}{\textbf{ASQA}} 
    & \multicolumn{2}{c}{\textbf{ELI5}} \\
\cmidrule(lr){2-3} \cmidrule(lr){4-5}
    & \textbf{Rouge-2} & \textbf{Rouge-L}
    & \textbf{Rouge-2} & \textbf{Rouge-L} \\
\midrule
Direct
    & 1.9 & 6.6
    & 1.1 & 7.5 \\
Native
    & 3.7 & 9.9
    & 1.8 & 10.0 \\
IRCoT
    & \underline{14.6} & \underline{30.8}
    & 4.9 & \textbf{21.6} \\
MetaRAG
    & 4.4 & 10.9
    & 1.1 & 7.3 \\
GenGround
    & 12.9 & 29.3
    & \underline{3.8} & \underline{18.2} \\
\textbf{DualRAG}
    & \textbf{15.4} & \textbf{31.7}
    & \underline{3.8} & 18.1 \\
\bottomrule
\end{tabular}
\caption{Evaluation results on long-format question answering datasets, using Qwen2.5-72B-Instruct. We evaluate using ROUGE~\cite{metrics/rouge}}
\label{tab:exp-long-format}
\end{table}

\subsection{Case Study}

We conduct case studies and find that DualRAG effectively identifies knowledge needs during reasoning, generates tailored queries, and organizes retrieved knowledge to facilitate high-quality answer generation. Detailed examples and analyses are provided in Appendix~\ref{appendix:case}.

\section{Conclusion}

We propose DualRAG, a dual-process RAG framework that tightly integrates Reasoning-augmented Querying (RaQ) and progressive Knowledge Aggregation (pKA) to address multi-hop QA problems.
RaQ dynamically identifies knowledge demands and formulates targeted retrieval queries, while pKA structures and refines acquired information to support coherent reasoning. Meanwhile, by fine-tuning DualRAG on smaller models, we ensure that it maintains its strong reasoning and retrieval capabilities, allowing it to retain its core advantages even with reduced resource consumption. Experimental results on multiple datasets demonstrate that DualRAG significantly improves answer accuracy and coherence, confirming its effectiveness as a robust and efficient solution for complex reasoning tasks.

\section*{Limitations}

Despite the effectiveness of our approach, several limitations remain.
First, while our method exhibits some robustness to noise, real-world retrieval corpora may still suffer from missing, contradictory, or outdated knowledge, which can impact reasoning reliability. To address these challenges, future improvements could involve enhancing the Knowledge Summarizer to better resolve inconsistencies and infer missing information.
Second, although our method reduces retrieval frequency by identifying knowledge needs, thereby minimizing interruptions to reasoning, it still introduces additional latency due to multi-turn retrieval. A promising direction for improvement is to enhance the Knowledge Outline, enabling it to progressively accumulate and reuse knowledge across questions, thereby reducing retrieval dependency on multi-turn retrieval and facilitating continuous learning and self-improvement.

\section*{Future work}

In future work, we plan to incorporate reinforcement learning to further enhance DualRAG’s performance. Reinforcement learning can enable the system to better optimize retrieval and reasoning strategies through continuous feedback, improving adaptability and efficiency in complex tasks.

We also aim to apply DualRAG in AI for Science, where structured retrieval and reasoning are crucial. For example, in biological sciences, frameworks like CellAgent \cite{xiao2024cellagent} highlight the potential of LLMs in automating scientific workflows. Similarly, in scientific domains such as software testing \cite{zheng2019wuji} and tokamak plasma control \cite{degrave2022magnetic}, DualRAG can complement deep reinforcement learning by organizing domain knowledge and guiding exploration.

\section*{Acknowledgments}

This work is supported by the National Key Research and Development Program of China (Grant No. 2024YFE0210900), the National Natural Science Foundation of China (Grant Nos. 62422605, 92370132) and the Xiaomi Young Talents Program of Xiaomi Foundation.

\bibliography{custom}

\appendix

\clearpage

\section{Prompts}

The prompts for Reasoner, Entity Identifier, and Knowledge Summarizer are shown in Figures~\ref{fig:prompt-reasoner},~\ref{fig:prompt-ei},~\ref{fig:prompt-ks}, respectively.

\newcounter{origsecnumdepth}
\setcounter{origsecnumdepth}{\value{secnumdepth}}
\setcounter{secnumdepth}{1}

\begin{figure*}[htbp]
    \begin{tcolorbox}[width=\textwidth, title={\textbf{Reasoner}}]
        \input{md/reasoner}
    \end{tcolorbox}
    \caption{Prompt for Reasoner}
    \label{fig:prompt-reasoner}
\end{figure*}

\begin{figure*}[htbp]
    \begin{tcolorbox}[width=\textwidth, title={\textbf{Entity Identifier}}]
        \input{md/ei}
    \end{tcolorbox}
    \caption{Prompt for Entity Identifier}
    \label{fig:prompt-ei}
\end{figure*}

\begin{figure*}[htbp]
    \begin{tcolorbox}[width=\textwidth, title={\textbf{Knowledge Summarizer}}]
        \input{md/ks}
    \end{tcolorbox}
    \caption{Prompt for Knowledge Summarizer}
    \label{fig:prompt-ks}
\end{figure*}

\setcounter{secnumdepth}{\value{origsecnumdepth}}

\section{Training Details}
\label{appdendix:training}

We use LLamaFactory~\cite{train/llamafactory} as the training framework and adopt DeepSpeed ZeRO Stage 3 optimization~\cite{train/deepseed} to enable efficient full-parameter fine-tuning. The learning rate is set to $3e^{-6}$, and we use a cosine learning rate scheduler. Experiments are conducted on 16 NVIDIA A100-PCIE-80GB GPUs, with a total training time of approximately 10 hours.

\section{Datasets Details}
\label{appaendix:datasets}

The details of the datasets used in this article are as follows:

\paragraph{NQ (Natural Questions)~\cite{benchmark/nq}}
NQ is a benchmark dataset for question answering research. It plays a crucial role in evaluating the ability of models to answer various types of questions. The dataset is sourced from Wikipedia, which provides a rich knowledge base. With 79,168 samples in the training set, 8,757 in the development set, and 3,610 in the test set, it offers a diverse range of questions and corresponding answers. These samples are designed to test the model's understanding of language, knowledge retrieval, and answer generation capabilities. By using NQ, researchers can assess how well their models perform in real - world - like question - answering scenarios.

\paragraph{PopQA~\cite{rag/later/trust}}
PopQA is a question - answering dataset that focuses on specific domains or popular knowledge. It draws its knowledge source from Wikipedia, leveraging the extensive information available there. Although the training set size is not specified, the development set contains 14,267 samples. This dataset is valuable for studying how models handle questions related to popular or specialized knowledge areas. It helps researchers understand the performance of models in retrieving and applying relevant knowledge from a well - known corpus like Wikipedia to answer questions within its scope.

\paragraph{ASQA~\cite{benchmark/asqa}}
ASQA aims to match factoid questions with long - form answers, contributing to the research of long - form question - answering tasks. It uses Wikipedia as its knowledge source, which enriches the dataset with reliable information. The training set of ASQA has 4,353 samples, and the development set has 948 samples. By providing such data, ASQA allows researchers to explore and develop models that can generate comprehensive and accurate long - form answers. It is useful for evaluating how well models can process and synthesize information to meet the requirements of long - form question - answering.

\paragraph{ELI5~\cite{benchmark/eli5}}
ELI5 is a long - form question - answering dataset based on the Reddit community. The questions and answers in this dataset typically revolve around daily life, scientific knowledge, and other common topics. With 272,634 samples in the training set and 1,507 in the development set, it offers a large - scale resource for studying detailed explanatory answers. Since it comes from a user - generated content platform, ELI5 reflects real - world language usage and the kind of questions people ask in an informal setting. This dataset helps researchers develop models that can generate natural - sounding and informative answers, similar to human - to - human explanations.

\paragraph{HotpotQA~\cite{benchmark/hotpotqa}}
HotpotQA is a dataset for diverse and explainable multi - hop question - answering tasks. It requires models to perform multiple reasoning steps to answer questions by integrating information from multiple documents, all sourced from Wikipedia. The training set consists of 90,447 samples, and the development set has 7,405 samples. By using HotpotQA, researchers can evaluate the multi - step reasoning ability of models. This dataset is crucial for understanding how well models can handle complex questions that demand the synthesis of information from different sources, and it promotes the development of more intelligent and explainable question - answering systems.

\paragraph{2WikiMultiHopQA~\cite{benchmark/2wikimultihop}}
2WikiMultiHopQA is specifically constructed to comprehensively evaluate the reasoning steps of models in multi - hop question - answering. It is based on Wikipedia knowledge, providing a solid foundation for multi - step reasoning tasks. The dataset has 15,000 samples in the training set and 12,576 in the development set. It enables researchers to study how models navigate through multiple pieces of information to answer complex questions. By analyzing the performance of models on 2WikiMultiHopQA, researchers can identify areas for improvement in multi - hop reasoning algorithms and enhance the overall quality of question - answering systems.

\paragraph{Musique~\cite{benchmark/musique}}
Musique is a multi - hop question - answering dataset focused on the music domain. It also uses Wikipedia as its knowledge source, enabling models to draw on a vast amount of music - related information. With 19,938 samples in the training set and 2,417 in the development set, Musique provides a platform for researchers to study how models perform multi - step reasoning in the music - specific context. This dataset helps in developing models that can answer complex music - related questions, which may require gathering and integrating information from multiple Wikipedia articles, and contributes to the advancement of domain - specific question - answering technology.

\section{Metrics Details}
\label{appendix:metric}

\paragraph{Exact Match (EM)}

\begin{align}
\text{EM}(y, \hat{y}) =
    \begin{cases} 
        1, & \text{if } \hat{y} = y \\
        0, & \text{otherwise}
    \end{cases}
\end{align}

\paragraph{Acc} Acc is more lenient than EM; it considers a match successful as long as the reference answer is a substring of the output.

\begin{align}
    \text{Acc}(y, \hat{y}) =
    \begin{cases} 
        1, & \text{if } y \text{ is a substring of } \hat{y} \\
        0, & \text{otherwise}
    \end{cases}
\end{align}

\paragraph{F1}

The F1 Score measures the token-level similarity between the output and the reference answer, with a value range of $ [0,1] $.

\begin{align}
    \text{F1}(y, \hat{y}) = \frac{2 \times P \times R}{P + R}
\end{align}

Where Precision ($P$) and Recall ($R$) are defined as follows, and these two metrics are also commonly used for evaluation:

\begin{align}
    P = \frac{| \text{CommonTokens}(y, \hat{y}) |}{| \text{Tokens}(\hat{y}) |} \\  
    R = \frac{| \text{CommonTokens}(y, \hat{y}) |}{| \text{Tokens}(y) |}
\end{align}

For example, for the question: \textit{Who is the mother of the director of the film Polish-Russian War (Film)?} the reference answer is "Małgorzata Braunek".

\begin{itemize}
    \item If the model outputs \textit{Małgorzata Braunek}, then $EM=1$, $ACC=1$, $F1=1$.
    \item If the model outputs \textit{The mother of the director of the film 'Polish-Russian War' is Małgorzata Braunek.}, then $EM$=0, $ACC$=1, and $F1$ is between 0 and 1, as the reference answer is part of the output.
\end{itemize}

\paragraph{$\text{Acc}^{\dagger}$}

LLM as a judge is widely adopted in many works~\cite{survey/eval}. We follow the evaluation methodology of \cite{rag/pipeline/iter-retgen}, adopting the same judge prompt. Qwen2.5-72B-Instruct~\cite{model/qwen2.5} is used as the evaluation model to assess the correctness of generated responses based on the question, the model-generated answer, and the ground truth.

\begin{figure*}[htbp]
    \begin{tcolorbox}[width=\textwidth, title={\textbf{Judger}}]
        \input{md/judge}
    \end{tcolorbox}
    \caption{Prompt for Judger}
    \label{fig:prompt-judge}
\end{figure*}

\section{Efficiency Analysis}

In our framework, the primary contributors to reasoning latency include:  
\textbf{(1)} the overhead of large model inference, which is largely determined by the number of generated tokens;  
\textbf{(2)} the number of retrieval triggers, i.e., the number of iterations performed.

To assess the computational efficiency of DualRAG, we conducted a quantitative analysis as follows.

\paragraph{Token Consumption Across Methods.}  
Table~\ref{tab:token-overhead} presents the \textbf{average output token overhead} per question for different methods. We observe that DualRAG maintains a similar token budget to existing iterative retrieval-based approaches, with the average consumption consistently below 1000 tokens.

\begin{table}[htbp]
\centering
\small
\setlength{\tabcolsep}{8pt}
\renewcommand{\arraystretch}{1.1}
\begin{tabular}{lcccc}
\toprule
\textbf{Method} & \textbf{HQA} & \textbf{WQA} & \textbf{MQA} & \textbf{Mean} \\
\midrule
GenGround   & 978.9 & 714.0 & 893.7 & 862.2 \\
MetaRAG     & 482.2 & 687.5 & 623.9 & 597.9 \\
\textbf{DualRAG} & \textbf{601.3} & \textbf{555.2} & \textbf{637.7} & \textbf{598.1} \\
\bottomrule
\end{tabular}
\caption{Average output token consumption per question across three datasets.}
\label{tab:token-overhead}
\end{table}

To better understand where computation is spent within DualRAG, Table~\ref{tab:token-breakdown} reports the \textbf{token usage of each major module}: inference (Infer), evidence identification (EI), and knowledge selection (KS). The sum of these components matches the total cost shown previously.

\begin{table}[htbp]
\centering
\small
\setlength{\tabcolsep}{8pt}
\renewcommand{\arraystretch}{1.1}
\begin{tabular}{lcccc}
\toprule
\textbf{Module} & \textbf{HQA} & \textbf{WQA} & \textbf{MQA} & \textbf{Mean} \\
\midrule
Infer & 196.2 & 203.1 & 226.6 & 208.7 \\
EI    & 192.5 & 187.1 & 239.5 & 206.4 \\
KS    & 212.6 & 165.0 & 171.6 & 183.0 \\
\textbf{Total} & \textbf{601.3} & \textbf{555.2} & \textbf{637.7} & \textbf{598.1} \\
\bottomrule
\end{tabular}
\caption{Token usage per module within DualRAG across datasets.}
\label{tab:token-breakdown}
\end{table}

\paragraph{Retrieval Trigger Frequency.}  
As shown in Figure~\ref{fig:exp/steps} of our main paper, we also analyze the number of retrievals triggered per question across methods. The results indicate that \textbf{DualRAG significantly reduces retrieval iterations}, with the vast majority of questions answered using no more than two retrieval rounds.

DualRAG demonstrates a strong balance between reasoning quality and computational efficiency. By minimizing unnecessary retrievals and managing token consumption across modules, it offers a practical solution for real-world deployment scenarios.

\section{Case Study}
\label{appendix:case}

We conducted several case studies to analyze the effectiveness of our method. The results show that DualRAG can dynamically retrieve information when additional knowledge is required, enabling efficient multi-hop reasoning with minimal iterations. Detailed case study examples are illustrated in Table~\ref{tab:case}.

\begin{table*}[htbp]
\centering
\small
\setlength{\tabcolsep}{2.5pt}
\begin{tabular}{p{0.95\linewidth}}

\toprule

\multicolumn{1}{c}{\textbf{Case Study}} \\

\midrule

\textbf{Question}: Which film has the director who was born later, El Extraño Viaje or Love In Pawn? \\
\textbf{Gold Answers}: [ El extraño viaje, Extraño viaje, El Extraño Viaje ] \\
\midrule
\textbf{Step 1} \\

\textbf{Reasoning-augmented Querying} \\

\textcolor{brown}{Reasoner}: To determine which film has the director who was born later, I need to find the birth years of the directors of El Extraño Viaje and Love In Pawn. \textcolor{red}{Reason interrupt for retrieval} \\

\textcolor{brown}{Entity Identifier}: \\
(1) El Extraño Viaje: [El Extraño Viaje director, El Extraño Viaje director birth year] \\
(2) Love In Pawn: [Love In Pawn director, When was the director of Love In Pawn born] \\
\\
\textbf{Retrieved Documents} \\

(1) El extraño viaje: El extraño viaje is a 1964 Spanish black drama film directed by Fernando Fernán Gómez... \\
(2) Love in Pawn: Love in Pawn is a 1953 British comedy film directed by Charles Saunders and starring Bernard Braden and Barbara Kelly... \\
(3) A Lover in Pawn: A Lover in Pawn is a 1920 Swedish silent drama film directed by Victor Sjöström... \\
(4) ... \\
\\
\textbf{progressive Knowledge Aggregation} \\

\textcolor{brown}{Knowledge Summarizer} \\

(1) El extraño viaje: El extraño viaje is a 1964 Spanish black drama film directed by Fernando Fernán Gómez. \\
(2) Love In Pawn: Love in Pawn is a 1953 British comedy film directed by Charles Saunders and starring Bernard Braden and Barbara Kelly. \\
\midrule
\textbf{Step 2} \\

\textbf{Reasoning-augmented Querying} \\

\textcolor{brown}{Reasoner}: I need to find the birth years of Fernando Fernán Gómez, the director of El Extraño Viaje, and Charles Saunders, the director of Love In Pawn. \textcolor{red}{Reason interrupt for retrieval} \\

\textcolor{brown}{Entity Identifier} \\
(1) Fernando Fernán Gómez: [Fernando Fernán Gómez, Fernando Fernán Gómez birth year] \\
(2) Charles Saunders: [Charles Saunders birth year] \\
\\
\textbf{Retrieved Documents} \\

(1) Fernando Fernán Gómez: Fernando Fernández Gómez (28 August 1921 – 21 November 2007) better known as... \\
(2) La vida alrededor: La vida alrededor is a 1959 Spanish comedy film written, starred... \\
(3) 7000 días juntos: 7000 días juntos is a 1994 film directed by... \\
(4) harles Saunders (director): Charles Joel Saunders (8 April 1904 – April 1997) was an English film director... \\
(5) ... \\
\\
\textbf{progressive Knowledge Aggregation} \\

\textcolor{brown}{Knowledge Summarizer} \\
(1) Fernando Fernán Gómez: Fernando Fernández Gómez, better known as Fernando Fernán-Gómez, was born on 28 August 1921. \\
(2) Charles Saunders: Charles Joel Saunders (8 April 1904 – April 1997) was an English film director and screenwriter. \\
\midrule
\textbf{Step 3} \\

\textbf{Reasoning-augmented Querying} \\
\textcolor{brown}{Reasoner}: From the known knowledge, Fernando Fernán Gómez was born on 28 August 1921, and Charles Saunders was born on 8 April 1904. Since 1921 is later than 1904, the director born later is Fernando Fernán Gómez, who directed El Extraño Viaje. \textcolor{blue}{Reasoning completed} \\
\bottomrule
\end{tabular}
\caption{Case Study}
\label{tab:case}
\end{table*}

\section{Error Analysis}

Although we designed a dual-process architecture to optimize multi-hop reasoning, due to the limitations of large model reasoning capabilities and the inherent flaws in retrieval tools, DualRAG may still fail on certain problems. \textbf{We manually analyzed the error cases in 100 test samples}, and after excluding issues caused by knowledge gaps in the retrieval corpus, we found that the errors primarily fall into the following two categories:

\paragraph{LLM reasoning hallucinations (75\%)} In some cases, the model may infer incorrect conclusions based on the documents retrieved.

For example, for the question \textit{When did John V, Prince of Anhalt-Zerbst's father die?}

After retrieving the following document: \textit{John V of Anhalt-Zerbst (Dessau, 4 September 1504 – Zerbst, 4 February 1551), was a German prince of the House of Ascania and ruler of the principality of Anhalt-Dessau. \textbf{From 1544}, he assumed rule of the re-created principality of Anhalt-Zerbst. John was the second (but eldest surviving) \textbf{son of Ernest I, Prince of Anhalt-Dessau}, by his wife Margarete, daughter of Henry I, Duke of Münsterberg-Oels, and granddaughter of George of Poděbrady, King of Bohemia.}

DualRAG correctly identifies \textit{Ernest I, Prince of Anhalt-Dessau} as \textit{John V's father}. However, due to the statement From \textit{1544}, he assumed rule of \textit{Anhalt-Zerbst}," the model incorrectly infers that \textit{John V}'s father died in the same year he assumed the throne, even though the document does not explicitly provide his father's death date. This requires further retrieval to confirm this information.

\paragraph{Misunderstanding of the question (25\%)} In some cases, the model may misunderstand the intent of the question, particularly when the question contains ambiguous terms (e.g., "country") or implicit references.

For example, for the question: \textit{What is the award that the performer of the song Sunday Papers earned?}

DualRAG incorrectly interprets the question as asking, "\textit{What award did the song Sunday Papers earn?}" when, in fact, the question is asking about the awards received by the performer of the song.

\section{More Discussions}

\subsection{The Necessity of Multi-Step Retrieval}

To evaluate the effectiveness of multi-turn retrieval, we conducted experiments on the HotpotQA dataset using GPT-4o-mini\footnote{\url{https://openai.com/index/gpt-4o-mini-advancing-cost-efficient-intelligence/}}. Specifically, we compared the following three retrieval strategies:

\begin{itemize}
    \item \textbf{NativeRAG-10}: A baseline setup consistent with prior experiments, retrieving up to 10 relevant documents in a single step.
    \item \textbf{NativeRAG-200}: A variant that retrieves up to 200 documents in one step to simulate broader retrieval coverage.
    \item \textbf{2HopRAG}: A two-hop approach. In the first step, 10 documents are retrieved using the original query. In the second step, each retrieved document is used as a sub-query to retrieve additional documents, resulting in up to 110 documents overall.
\end{itemize}

Table~\ref{tab:multi-hop-retrieval} presents the comparative results of these strategies.

\begin{table}[htbp]
\centering
\small
\setlength{\tabcolsep}{8pt}
\renewcommand{\arraystretch}{1.1}
\begin{tabular}{lcccc}
\toprule
\textbf{Method} & \textbf{$\text{Acc}^{\dagger}$} & \textbf{EM} & \textbf{ACC} & \textbf{F1} \\
\midrule
NativeRAG-10  & 0.553 & 0.333 & 0.475 & 0.523 \\
NativeRAG-200 & 0.550 & 0.333 & 0.475 & 0.487 \\
\textbf{2HopRAG}     & \textbf{0.681} & \textbf{0.444} & \textbf{0.603} & \textbf{0.575} \\
\bottomrule
\end{tabular}
\caption{Performance comparison of different retrieval strategies on HotpotQA using GPT-4o-mini.}
\label{tab:multi-hop-retrieval}
\end{table}

Although \textit{NativeRAG-200} expands the retrieval scope significantly, it does not lead to noticeable performance improvement over \textit{NativeRAG-10}. In contrast, the two-hop strategy of \textit{2HopRAG} yields substantial gains across all metrics.

This demonstrates that:

\begin{itemize}
    \item Multi-hop reasoning often requires the gradual construction of an \textit{information chain}.
    \item One-step retrieval, regardless of breadth, struggles to fully capture the sequential dependencies inherent in such tasks.
\end{itemize}

These results validate the necessity of \textbf{multi-step retrieval} for complex reasoning tasks. A multi-turn approach not only enhances coverage but also aligns better with the step-wise nature of human-like reasoning.

\subsection{Iteration Scaling}
\label{app:scaling}

Existing methods often treat retrieval as a fixed process, either based on static trigger conditions or by first generating heuristic responses and then deciding whether to initiate further retrievals based on the quality of the output. However, such approaches lack a forward-looking understanding of knowledge requirements. This limitation makes it challenging to dynamically adjust the timing and frequency of retrievals, potentially introducing redundant information or causing delays that disrupt the reasoning process.

To quantify this difference, we analyzed the average number of retrieval iterations required by various IterativeRAG methods, as shown in Figure~\ref{fig:exp/steps} of our paper. We summarize the key findings in the table below:

\begin{table}[htbp]
\centering
\small
\begin{tabular}{lccc}
\toprule
\textbf{Method} & \textbf{HQA} & \textbf{WQA} & \textbf{MQA} \\
\midrule
IRCoT & 3.76 & 4.19 & 4.26 \\
MetaRAG & 2.93 & 3.14 & 3.12 \\
\textbf{DualRAG} & \textbf{1.64} & \textbf{1.99} & \textbf{2.20} \\
\bottomrule
\end{tabular}
\caption{Average number of retrieval iterations required by different methods.}
\label{tab:retrieval-frequency}
\end{table}

To further demonstrate the efficiency of our approach, we conducted experiments on the 2WikiMultiHopQA dataset, systematically varying the maximum number of allowed retrieval iterations. The results are summarized in Table~\ref{tab:iteration-scaling} and illustrate that our method (RaQ) can achieve strong performance with minimal retrieval overhead. Notably, only two iterations are sufficient to solve most queries effectively. This conclusion is consistent with the statistical trends shown in Figure~\ref{fig:exp/steps}, which highlights the reasoning depth across different methods.

\begin{table}[htbp]
\centering
\small
\setlength{\tabcolsep}{2pt}
\begin{tabular}{lcccccc}
\toprule
\textbf{Max Iterations} & 0 & 1 & 2 & 3 & 4 & 5 \\
\midrule
\textbf{F1 Score} & 0.385 & 0.556 & \textbf{0.736} & 0.733 & 0.733 & 0.722 \\
\bottomrule
\end{tabular}
\caption{Impact of limiting the maximum number of retrieval iterations on performance (F1 score) for 2WikiMultiHopQA.}
\label{tab:iteration-scaling}
\end{table}

We also extended this analysis to other baseline methods. As shown in Table~\ref{tab:baseline-scaling}, most competing methods require more iterations to reach comparable performance. For example, only GenGround achieves a similar score to ours after eight iterations, whereas DualRAG reaches that level after just two. Other baselines still show a significant performance gap even after more iterations.

\begin{table*}[htbp]
\centering
\small
\setlength{\tabcolsep}{4pt}
\begin{tabular}{lcccccccc}
\toprule
\multirow{2}{*}{\textbf{Max Iterations}} & \multicolumn{2}{c}{\textbf{2}} & \multicolumn{2}{c}{\textbf{3}} & \multicolumn{2}{c}{\textbf{5}} & \multicolumn{2}{c}{\textbf{8}} \\
\cmidrule(lr){2-3} \cmidrule(lr){4-5} \cmidrule(lr){6-7} \cmidrule(lr){8-9}
& Acc & F1 & Acc & F1 & Acc & F1 & Acc & F1 \\
\midrule
IRCoT     & 66.9 & 51.9 & 78.4 & 60.7 & 83.6 & 67.5 & 85.5 & 66.9 \\
GenGround & 79.8 & 67.5 & 81.3 & 67.2 & 78.1 & 70.3 & 82.4 & 74.0 \\
MetaRAG   & 55.3 & 56.3 & 56.3 & 56.9 & 59.7 & 61.2 & 61.6 & 60.9 \\
\textbf{DualRAG} & \textbf{83.5} & \textbf{75.6} & \textbf{83.6} & \textbf{75.9} & \textbf{85.0} & \textbf{77.3} & \textbf{84.3} & \textbf{76.4} \\
\bottomrule
\end{tabular}
\caption{Performance comparison of different methods under varying maximum iteration constraints.}
\label{tab:baseline-scaling}
\end{table*}

These findings demonstrate that our framework, by \textbf{proactively identifying knowledge demands}, minimizes the number of retrievals—i.e., interruptions to the reasoning process—to its theoretical minimum. In the datasets used, most questions require 1–2 reasoning hops, though some involve up to 4–5 documents. DualRAG closely aligns retrieval steps with this minimal number, reflecting its ability to avoid unnecessary trial-and-error iterations. Rather than relying on simple iterative optimization, DualRAG offers a more efficient and effective retrieval strategy.

\subsection{Enhanced Variant of \textit{w/o KO} Ablation}

To further assess the importance of the \textbf{Knowledge Outline (KO)} mechanism in our framework, we conducted an enhanced variant ablation, denoted as \textit{w/o KO$^\ast$}.

While the original \textit{w/o KO} setting removes the KO module entirely and directly feeds raw retrieved documents into the Reasoner, the \textit{w/o KO$^\ast$} variant retains the summarization step but skips the construction of a structured knowledge outline. This experiment simulates a setting where the model must reason based on flat summaries, without the benefit of cross-document structuring provided by KO.

The results of \textit{w/o KO$^\ast$} are presented in Table~\ref{tab:appendix-ko-star}, showing a consistent decline in performance across all benchmarks:

\begin{table}[htbp]
\centering
\small
\setlength{\tabcolsep}{4.5pt}
\begin{tabular}{lcccccc}
\toprule
\multirow{2}{*}{\textbf{Method}}
    & \multicolumn{2}{c}{\textbf{HQA}}
    & \multicolumn{2}{c}{\textbf{WQA}}
    & \multicolumn{2}{c}{\textbf{MQA}} \\
\cmidrule(lr){2-3} \cmidrule(lr){4-5} \cmidrule(lr){6-7}
    & \textbf{Acc} & \textbf{F1}
    & \textbf{Acc} & \textbf{F1}
    & \textbf{Acc} & \textbf{F1} \\
\midrule
\textbf{DualRAG}
    & 70.0 & 65.7
    & 85.0 & 77.3
    & 66.3 & 56.3 \\
- w/o KO
    & 69.5 & 64.0
    & 79.4 & 72.8
    & 58.1 & 49.8 \\
- w/o KO$^\ast$
    & 69.7 & 65.1
    & 74.3 & 66.2
    & 55.8 & 49.2 \\
\bottomrule
\end{tabular}
\caption{Performance of the enhanced \textit{w/o KO} variant where the Knowledge Outline is removed, but summarization is retained.}
\label{tab:appendix-ko-star}
\end{table}

These results demonstrate that while summarization alone does offer some level of abstraction, it falls short of the benefits provided by a structured Knowledge Outline. Without KO, the model struggles to effectively organize and leverage cross-document information, which not only impairs reasoning quality but may also lead to inefficiencies—such as processing overly verbose or less relevant content. This highlights the importance of KO as both a cognitive aid for the LLM and a practical design for scalable reasoning.

\subsection{Ablation on the Reranker}

To assess the impact of the reranker module in our framework, we conducted an ablation study. Specifically, we modified the original retrieval setup—which first retrieves the top-50 candidates and then applies a reranker to select the top-10 (as described in lines 470–480)—to directly retrieve the top-10 documents without reranking. The comparative results are shown in Table~\ref{tab:reranker-ablation}.

\begin{table}[htbp]
\centering
\small
\setlength{\tabcolsep}{4.5pt}
\begin{tabular}{lcccccc}
\toprule
\multirow{2}{*}{\textbf{Method}} & \multicolumn{2}{c}{\textbf{HQA}} & \multicolumn{2}{c}{\textbf{WQA}} & \multicolumn{2}{c}{\textbf{MQA}} \\
\cmidrule(lr){2-3} \cmidrule(lr){4-5} \cmidrule(lr){6-7}
 & Acc & F1 & Acc & F1 & Acc & F1 \\
\midrule
DualRAG          & 70.0 & 65.7 & 85.0 & 77.3 & 66.3 & 56.3 \\
- w/o rerank & 67.6 & 63.3 & 86.3 & 76.4 & 64.8 & 54.9 \\
\bottomrule
\end{tabular}
\caption{Ablation results on the reranker component across three datasets.}
\label{tab:reranker-ablation}
\end{table}

As shown, removing the reranker module does lead to a slight drop in performance, though the decrease is not as substantial as that observed when ablating other core components of our system (see Table~\ref{tab:exp-abl-framework}). This observation highlights the robustness of our design.

The reranker serves as a more precise scoring model during retrieval. The relatively modest performance degradation observed in this ablation suggests that the dual-process structure of DualRAG is capable of partially compensating for limitations in retrieval quality. In other words, our design enables the reasoning capabilities of large language models (LLMs) to better utilize—and even enhance—the utility of standard retrieval tools, thereby narrowing the gap with stronger retrieval systems.

We would also like to emphasize two additional points:

\begin{itemize}
    \item \textbf{The reranker is not redundant}: It still contributes meaningfully to retrieval quality. However, thanks to our structured and adaptive framework, even with a simpler retriever, comparable results can be achieved.
    \item \textbf{Fair comparison across baselines}: All models in our experiments, including DualRAG and the baselines, use the same retrieval settings—including reranker integration where applicable—ensuring a fair and consistent basis for comparison.
\end{itemize}

\subsection{More Baselines}

Methods such as FLARE~\cite{rag/self/flare} and DRAGIN~\cite{rag/self/dragin} dynamically decide \emph{when} to perform retrieval and \emph{which} tokens to query based on log-probability signals. While sharing similar objectives with our method—namely, improving the timing and targeting of retrieval—there are fundamental differences in implementation. FLARE relies on the model’s output uncertainty to trigger retrieval, whereas our method proactively anticipates and activates retrieval based on predicted knowledge demands, thereby enabling more structured and efficient knowledge usage.

Nonetheless, FLARE represents a strong and relevant baseline. We plan to include a more detailed discussion of such approaches in the revised version of our paper.

To further validate the effectiveness of our method, we conducted additional experiments using GPT-4o-mini. The results are shown in Table~\ref{tab:more-baselines}.

\begin{table*}[htbp]
\centering
\small
\setlength{\tabcolsep}{5.5pt}
\renewcommand{\arraystretch}{1.1}
\begin{tabular}{lcccccccccccc}
\toprule
\multirow{2}{*}{\textbf{Method}} 
    & \multicolumn{4}{c}{\textbf{HotpotQA}} 
    & \multicolumn{4}{c}{\textbf{2Wikimultihopqa}} 
    & \multicolumn{4}{c}{\textbf{MuSiQue}} \\
\cmidrule(lr){2-5} \cmidrule(lr){6-9} \cmidrule(lr){10-13}
    & \textbf{$\text{Acc}^{\dagger}$} & \textbf{EM} & \textbf{Acc} & \textbf{F1}
    & \textbf{$\text{Acc}^{\dagger}$} & \textbf{EM} & \textbf{Acc} & \textbf{F1}
    & \textbf{$\text{Acc}^{\dagger}$} & \textbf{EM} & \textbf{Acc} & \textbf{F1} \\
\midrule
FLARE   & 0.728 & 0.427 & 0.661 & 0.573 
        & 0.687 & 0.504 & 0.772 & 0.632 
        & 0.665 & 0.355 & 0.628 & 0.502 \\
\textbf{DualRAG} 
        & \textbf{0.767} & \textbf{0.468} & \textbf{0.679} & \textbf{0.617} 
        & \textbf{0.784} & \textbf{0.612} & \textbf{0.814} & \textbf{0.722} 
        & \textbf{0.659} & \textbf{0.383} & \textbf{0.611} & \textbf{0.528} \\
\bottomrule
\end{tabular}
\caption{Comparison of DualRAG and FLARE on three MHQA datasets using GPT-4o-mini.}
\label{tab:more-baselines}
\end{table*}

\subsection{Other LLMs}

We chose the Qwen series models for our experiments based on the following considerations:

\begin{itemize}
  \item \textbf{Strong natural language processing capabilities.} The Qwen series models have demonstrated excellent performance across a wide range of natural language processing tasks, with outstanding knowledge integration and reasoning abilities.
  \item \textbf{Open-source reproducibility.} The Qwen series models are open-source, and the community provides stable API support, which ensures high reproducibility of our experiments and facilitates further research and validation by the community.
\end{itemize}

\textbf{DualRAG does not rely on a specific LLM architecture}, as the LLM, serving as the core "brain" of DualRAG, offers high scalability and flexibility, making it not limited to any particular model.

To verify the applicability of our approach across different LLM architectures, we also \textbf{conducted experiments on other open-source models} (Llama-3.1-70B-Instruct\footnote{\url{https://huggingface.co/meta-llama/Meta-Llama-3-70B-Instruct}} and Mistral-Small-24B-Instruct-2501\footnote{\url{https://huggingface.co/mistralai/Mistral-Small-24B-Instruct-2501}}). The results are shown in Table~\ref{tab:more-llm}.

\begin{table*}[htbp]
\centering
\small
\setlength{\tabcolsep}{5.5pt}
\renewcommand{\arraystretch}{1.1}
\begin{tabular}{lcccccccccccc}
\toprule
\multirow{2}{*}{\textbf{Method}} 
    & \multicolumn{4}{c}{\textbf{HotpotQA}} 
    & \multicolumn{4}{c}{\textbf{2Wikimultihopqa}} 
    & \multicolumn{4}{c}{\textbf{MuSiQue}} \\
\cmidrule(lr){2-5} \cmidrule(lr){6-9} \cmidrule(lr){10-13}
    & \textbf{$\text{Acc}^{\dagger}$} & \textbf{EM} & \textbf{Acc} & \textbf{F1}
    & \textbf{$\text{Acc}^{\dagger}$} & \textbf{EM} & \textbf{Acc} & \textbf{F1}
    & \textbf{$\text{Acc}^{\dagger}$} & \textbf{EM} & \textbf{Acc} & \textbf{F1} \\
\midrule
\rowcolor[gray]{0.9} \textit{Base LLM} & \multicolumn{12}{c}{\textit{LLaMA-3.1-70B-Instruct}} \\
Direct    & 0.546 & 0.267 & 0.364 & 0.415 & 0.388 & 0.254 & 0.349 & 0.350 & 0.306 & 0.091 & 0.191 & 0.199 \\
Native    & 0.728 & 0.478 & 0.534 & 0.610 & 0.479 & 0.370 & 0.389 & 0.422 & 0.408 & 0.260 & 0.301 & 0.342 \\
IRCoT     & 0.816 & 0.528 & 0.710 & 0.654 & 0.733 & 0.557 & 0.790 & 0.644 & 0.606 & 0.338 & 0.522 & 0.457 \\
GenGround & 0.835 & 0.514 & 0.713 & 0.665 & 0.843 & 0.692 & 0.862 & 0.781 & 0.593 & 0.409 & 0.563 & 0.492 \\
DualRAG   & \textbf{0.834} & \textbf{0.522} & \textbf{0.732} & \textbf{0.675} & \textbf{0.890} & \textbf{0.730} & \textbf{0.887} & \textbf{0.815} & \textbf{0.684} & \textbf{0.405} & \textbf{0.634} & \textbf{0.537} \\
\midrule
\rowcolor[gray]{0.9} \textit{Base LLM} & \multicolumn{12}{c}{\textit{Mistral-Small-24B-Instruct-2501}} \\
Direct    & 0.367 & 0.229 & 0.246 & 0.323 & 0.322 & 0.277 & 0.291 & 0.326 & 0.161 & 0.074 & 0.096 & 0.148 \\
Native    & 0.646 & 0.436 & 0.475 & 0.572 & 0.411 & 0.315 & 0.348 & 0.384 & 0.367 & 0.235 & 0.274 & 0.339 \\
IRCoT     & 0.618 & 0.403 & 0.584 & 0.527 & 0.419 & 0.310 & 0.645 & 0.396 & 0.288 & 0.172 & 0.264 & 0.251 \\
GenGround & 0.726 & 0.421 & 0.640 & 0.572 & 0.710 & 0.510 & 0.765 & 0.618 & 0.578 & 0.298 & 0.538 & 0.430 \\
DualRAG   & \textbf{0.744} & \textbf{0.441} & \textbf{0.667} & \textbf{0.604} & \textbf{0.761} & \textbf{0.549} & \textbf{0.785} & \textbf{0.650} & \textbf{0.601} & \textbf{0.310} & \textbf{0.567} & \textbf{0.461} \\
\bottomrule
\end{tabular}
\caption{Evaluation results of DualRAG and baselines on three MHQA datasets using two different open-source LLMs. Bold indicates best performance.}
\label{tab:more-llm}
\end{table*}

These experimental results indicate that our method is also applicable to other LLM architectures, demonstrating robust performance across different models.

\end{document}